\documentclass[11pt]{article}

\PassOptionsToPackage{table}{xcolor}
\usepackage[preprint]{acl}

\usepackage{times}
\usepackage{latexsym}
\usepackage[T1]{fontenc}
\usepackage{iftex}
\ifPDFTeX\usepackage[utf8]{inputenc}\fi
\usepackage{microtype}
\usepackage{inconsolata}

\usepackage{graphicx}
\usepackage{booktabs}
\usepackage{xcolor}
\usepackage{amsmath}
\usepackage[capitalise,noabbrev]{cleveref}
\usepackage{tikz}
\usetikzlibrary{positioning,arrows.meta,calc}
\usepackage{fontawesome5}
\usepackage{placeins}
\graphicspath{{figures/}}

\title{Language Models Make New Research Sound Old}

\author{Guy Zyskind \\ University of Miami}

\begin{document}
\maketitle

\begin{abstract}
Researchers are increasingly using large language models (LLMs) to validate research questions. We find that LLMs make open research questions sound partly studied and overstate what related papers establish. Models recognize most of these overstatements when asked to check the papers' full texts separately. On 40 questions drawn from newly accepted papers, GPT-6-Astra correctly judges 98\% to be unstudied, but states that status plainly in only 8\% of ordinary answers. All six frontier systems show reporting gaps of 30--90 percentage points. Across 22 models given identical papers, higher capability is associated with more accurate judgments and larger reporting gaps. This reporting failure is asymmetric: models correctly identify studied questions in 88--100\% of ordinary answers when we supply the answering papers. To correct this, we build a fixed answer from each model's judgment and separate checks of cited papers. This raises Astra's clear reporting of open status from 8\% to 95\%, with improvements in all six systems. Finally, we predict reviewers' novelty grades by combining several models' written ratings with a rule learned from graded examples, reaching the performance range reported by Think-Probe-Respond. Unlike TPR, our approach needs no access to model weights or internal activations, so it also works with closed-weight frontier models.
\end{abstract}

\section{Introduction}
\label{sec:intro}
\begin{figure}[t]
% Accepted-paper item PN-25 (CIDER, COLM 2026, OpenReview 6GTN8QkGmY).
% Question: followups/peer_novel/items/items.jsonl (shortened below).
% Normal and forced replies: followups/peer_novel/more_models/runs/
%   gpt6sol_normal/PN-25.json and gpt6sol_forced/PN-25.json.
% Both use the normal reply's retrieved evidence, reference date 2026-03-27.
% Claim: uid ff0eb52050d81f, both coders a=3, independent full-text grade=2.
%   followups/peer_novel/more_models/grades/claim_grades.jsonl
% Own full-text grade=2: followups/selfgrade_main/data/selfgrades.jsonl.
% Acceptance and date checks: followups/peer_novel/data/colm2026_accepted.json
%   and logs/date_filter.jsonl. The target paper is blocked from search.
\centering
\begingroup
\definecolor{teaserBlue}{HTML}{0072B2}
\definecolor{teaserOrange}{HTML}{E69F00}
\begin{tikzpicture}[
  font=\footnotesize,
  bubble/.style={rounded corners=5pt, inner sep=5pt, align=flush left,
    text width=\dimexpr\columnwidth-12pt\relax}
]
\node[bubble,fill=black!6,anchor=north west] (question) at (0,0) {
  \faUserGraduate\ Has this question been studied?\\[3pt]
  Can a language model predict which personal information someone will share in a conversation from a handful of their past decisions, more accurately than without that history?
};
\node[bubble,fill=teaserOrange!13,anchor=north west] (answer) at ($(question.south west)+(0,-0.18)$) {
  \faRobot\ GPT-6-Sol's normal answer\\[3pt]
  ``\textbf{Yes---substantially}, though your exact setting may still be open.''\\[3pt]
  It cites \emph{Text-Based Personas for Simulating User Privacy Decisions} and calls the broad claim that past decisions improve privacy predictions ``difficult to present as new''.
};
\node[bubble,fill=teaserBlue!9,anchor=north west] (judgment) at ($(answer.south west)+(0,-0.18)$) {
  \textbf{Separate judgment, same model and evidence}\\[3pt]
  \textbf{NO (unstudied)}. Probability of prior study: 30\%.
};
\node[bubble,fill=black!6,anchor=north west] (paper) at ($(judgment.south west)+(0,-0.18)$) {
  \textbf{What the cited paper actually tested}\\[3pt]
  Several key parts differ (grade 2 of 4 for how closely the paper answers the question). The study tests other privacy settings and uses much more user history. The model agrees after reading the full text.
};
\end{tikzpicture}
\endgroup
\caption{The model describes an open question as substantially studied, although its separate judgment says unstudied. The question is shortened from a COLM 2026 paper published after training and excluded from search.}
\label{fig:teaser}
\end{figure}

As AI use in science grows, researchers often ask a language model whether a question they want to pursue has already been studied. \textcolor{black}{Reviewers face the same question when judging what a submission adds, and some already use language models to help write their reports \citep{liang2024monitoring}.} If a model overstates the overlap with earlier work, a researcher may set aside an open question or a reviewer may undervalue the paper that answers it. Overlooking an answering study can lead both to overestimate what is new.

We test whether models say clearly when a question is still open and whether the papers they cite cover as much of it as they claim. We find that seemingly thoughtful answers make new research sound old and overstate what related papers establish. The same models can make the right judgment but fail to communicate it: they identify questions as open when asked separately for a YES or NO judgement, yet obscure that conclusion in ordinary answers.

This gap appears across all six frontier systems we tested and reaches 90 percentage points for GPT-6-Astra: it correctly judges 98\% of the questions to be unstudied but states that status plainly in only 8\% of ordinary answers. We measured this on 40 questions from newly accepted papers at the Conference on Language Modeling (COLM). The papers appeared after the models' training, and peer review provided evidence that the questions were new. Figure~\ref{fig:teaser} illustrates this mismatch between judgment and advice.

We also asked whether this gap is smaller in more capable models and found the opposite. Across 22 open and closed models, each given the same research questions and abstracts of related papers, higher capability is associated with more accurate judgments and larger reporting gaps.

This reporting failure is asymmetric: models are much clearer when a question has already been studied, correctly reporting prior study in 88--100\% of ordinary answers across all six systems.

Models hedge more often on open questions when closely related work is available, suggesting that much of the proposed research has already been done. Replacing the closest paper, which leaves a required part unanswered, with one sharing only the topic reduces this hedging by 28 percentage points across the six systems.

How researchers ask their questions also matters. Among the five phrasings tested with Astra and Sonnet, the largest reporting gaps occur when we ask directly, ``Is this research question novel?''

An open question can still make only a small contribution, so we also examine reviewers' novelty grades. In recent work, Think-Probe-Respond (TPR) predicts these grades by looking inside open-weight models as they evaluate an idea \citep{schopf2026think}. We ask several current models for novelty ratings, average their written estimates, and learn from examples graded by reviewers how to convert the average into a final grade. The resulting predictor reaches TPR's reported performance range using ordinary API calls. It works with both closed and open models without access to their weights or internal states.

Our main contributions are:
\begin{itemize}
\item We identify an asymmetric reporting failure: models often recognize that a question is open in direct judgments but fail to say so in ordinary answers. Higher capability is associated with more accurate judgments and larger reporting gaps.
\item Models overstate how closely cited papers address a question, although they recognize most of these errors when checking the papers separately.
\item To simulate how researchers might interpret these answers, we use other AI models as readers deciding whether to pursue the project. These readers are almost four times as likely to call for an explanation of what is new after a hedged answer (84\% versus 23\%), even though all questions are open.
\item We reduce the reporting gap in all six systems using their own judgments and paper checks, raising Astra's clear reporting of open questions from 8\% to 95\%. Separately, we build a predictor of reviewers' novelty grades using only written model ratings. It reaches TPR's reported performance range without access to weights or internal states.
\end{itemize}

\section{Related Work}
\label{sec:related}

Existing novelty evaluations ask whether models agree with expert reviewers about what a contribution adds \citep{schopf2026novel,wu2026novbench}, with related literature providing evidence for those assessments \citep{shahid2025literature}. Instead, we approach the task primarily from the researcher's perspective, finding a large gap between what models say about a proposed research question in ordinary answers and what they judge when asked directly. Our separate comparison with TPR addresses novelty grading using written ratings in place of the internal model states required by its method \citep{schopf2026think}.

Earlier work on factual questions finds that models recognize errors in their own answers \citep{kadavath2022know} and that their internal states reveal correct answers they rarely produce \citep{orgad2025know,gekhman2025insideout}. Their wording also often misrepresents their confidence \citep{yona2024faithful}. In this work, we find that the failure in literature assistance is asymmetric. Frontier systems judge both studied and open questions accurately, yet their ordinary answers readily acknowledge prior study while obscuring what remains open. The reporting gap tends to be larger in more capable models even when they receive the same papers, despite their more accurate judgments.

The account of prior work in these answers raises a further concern. Studies of relevance and citation support find that models overrate related documents \citep{yu2026inflate} and attribute claims to sources that fail to support them \citep{liu2023verifiability,venkit2026deeptrace}. In novelty assessment, a model dismissed a new combination after finding its components in separate papers \citep{sternlicht2026old}, while NovGauge finds unsupported explanations even for correct judgments about overlap \citep{zhang2026novgauge}. Our question is instead whether assistants overstate how much of the researcher's proposed study a cited paper has already done. Independent checks find that papers described as nearly answering a question often leave several key parts unanswered, and the assistants recognize most of these mistakes when checking the papers separately. Replacing only the closest paper also reduces hedging about the same open questions, although models usually judge them open with either paper.

Human studies show that explanations and expressions of uncertainty affect trust in model answers \citep{steyvers2025know,kim2024sure,zhou2024relying}. Following work that evaluates answers through readers' interpretations \citep{band2024linguistic,stengeleskin2024lacie}, we ask other AI models whether they would pursue a project after reading the assistant's advice. These simulated readers usually recognize that the question remains open, but after hedged answers they more often advise explaining what is new. Such advice could discourage researchers from pursuing an open question. Our correction addresses the ambiguity by building the answer from the assistant's own judgment and checks of cited papers.

\section{Questions and Setup}
\label{sec:setup}

\textcolor{black}{For each question, we collect a \emph{normal answer} to the researcher's request and a \emph{forced answer} \textcolor{black}{with judgment $J\in\{\mathrm{YES},\mathrm{NO}\}$, where YES means studied and NO means not studied}. The forced answer also gives a probability of prior study. Each reply begins in a fresh session to avoid bias.}

\textcolor{black}{We label each normal answer $C\in\{\mathrm{studied},\mathrm{hedged},\mathrm{open}\}$ according to how it describes study status throughout the response. $\mathrm{studied}$ says the question has been studied. $\mathrm{open}$ states that it has not been studied and describes prior work consistently with that judgment. $\mathrm{hedged}$ covers the remaining answers: they call the question partly studied, call it open while describing related work as having done most of it, or leave its status otherwise unclear.} Two AI models, one from OpenAI and one from Anthropic, independently \textcolor{black}{assign $C$ to each answer}. They are not told which system wrote it or whether the question has actually been studied.

\textcolor{black}{We define the \emph{reporting gap} as the NO share of forced answers minus the share of $\mathrm{open}$ answers, in percentage points:}
{\color{black}
\[
\Delta=100\bigl[\Pr(J=\mathrm{NO})-\Pr(C=\mathrm{open})\bigr].
\]
}

\textcolor{black}{To test questions that were open before publication, we built a set of 40 \emph{accepted-paper questions} from COLM 2026 papers that appeared after the models' training.} Reviewers accepted these papers as new, providing ground-truth evidence for the open labels. We restricted each model's arXiv search to papers available before the submission deadline and excluded the answering paper. For the separate YES/NO judgment, we give each model the papers it retrieved for its normal answer, so both responses use the same evidence.

To compare systems without differences in their search results, we run a second test using a shared paper list for each question (supplied by Astra).

To test whether models distinguish studied from open questions, we build a controlled dataset of \emph{ladders} from published papers in LitSearch \citep{ajith2024litsearch}. Each ladder starts with a question answered by a published paper (L0). We create three versions of that question: change one core part (L1), change two (L2), or move it to a distant field (L3). A core part is the task or setting, method, outcome, or study design. We retain modified questions only when searches find no answering paper. We build the third dataset, \textcolor{black}{\emph{entity pairs}}, from questions with confirmed answering papers, replacing each real entity with an invented counterpart whose name returns no hits in scholarly or web searches.

Together, the ladders and entity pairs contain 503 questions. We compare normal answers with separate judgments across all six systems on a 100-question subset. This subset covers every question type and limits the cost of repeated searches and answer labeling. Astra, Luna and Sonnet also give normal answers on all 503 questions to provide more cited-paper claims at each ladder level. Each normal answer and separate judgment runs its own web search.

To test whether more capable models have smaller reporting gaps, we compare 22 models on a subset of 40 studied and 40 open questions from the ladders, plus the 40 \textcolor{black}{accepted-paper questions}. Every model receives the same papers for each question without searching.

Unless mentioned otherwise, experiments use OpenAI's Astra, Sol and Luna, and Anthropic's Sonnet, Opus and Fable (Appendix Table~\ref{tab:model_list}).

To test novelty, we use RINoBench \citep{schopf2026novel}, a dataset of research ideas from ICLR 2022 and 2023 submissions paired with reviewers' novelty grades from 1 to 5. We compare with TPR on this benchmark in Section~\ref{sec:tpr} \citep{schopf2026think}.

\emph{Notation and conventions.} \textcolor{black}{For comparisons with reviewer novelty grades, we use the ordered answer score $r(\mathrm{studied})=0$, $r(\mathrm{hedged})=1$, $r(\mathrm{open})=2$. Higher values mean the answer presents the question as more clearly open.} Brackets show 95\% bootstrap confidence intervals. We describe dataset construction in Appendix~\ref{app:sets} and scoring in Appendix~\ref{app:coding}.

\begin{figure*}[t]
\centering
\includegraphics[width=\textwidth]{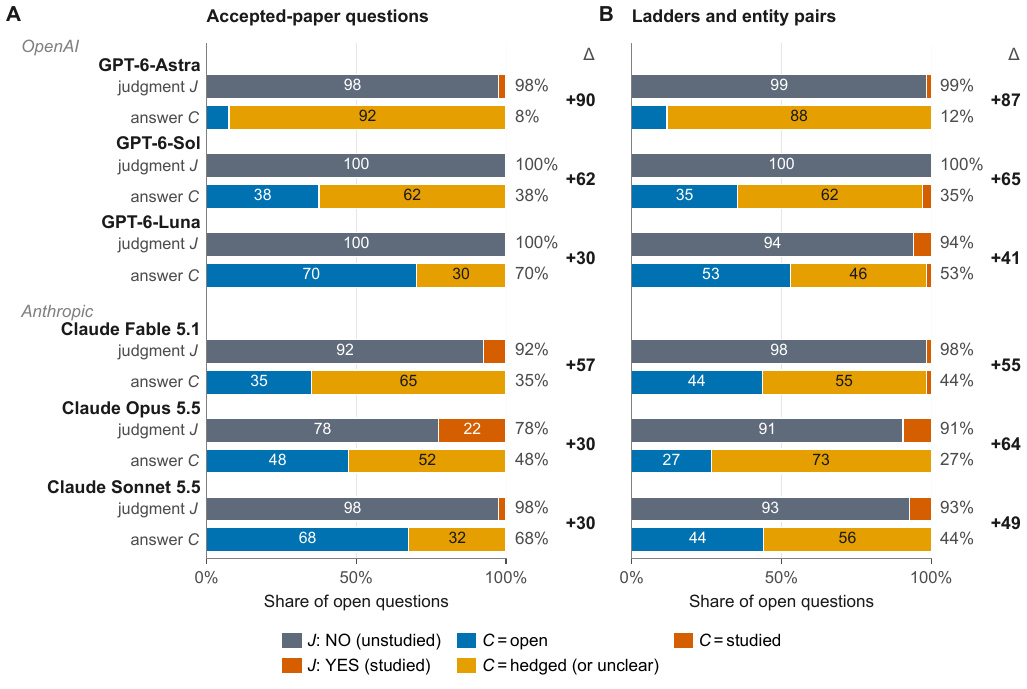}
\caption{\textcolor{black}{Models recognize that a question is open in separate judgments, yet often fail to say so clearly in ordinary answers. A shows accepted-paper questions and B over ladders and entity pairs dataset with live search. Upper bars show judgments $J$ and lower bars show answer labels $C$. Right columns report $J=\mathrm{NO}$, $C=\mathrm{open}$ and the reporting gap $\Delta$, using paired answers.}}
\label{fig:headline}
\end{figure*}

\section{Results}
\label{sec:results}

\subsection{Models identify open questions but do not say so plainly}
\label{sec:gap}

All six systems show reporting gaps of 30 to 90 points on the 40 questions drawn from newly accepted papers (Table~\ref{tab:verdicts}). Astra shows the largest gap: its separate judgments correctly identify 98\% as unstudied \textcolor{black}{($J=\mathrm{NO}$)}, yet only 8\% of its normal answers state this plainly \textcolor{black}{($C=\mathrm{open}$)}, a gap of 90 points [80, 98] (Figure~\ref{fig:headline}A).

Normal answers rarely say that the question has already been studied. Instead, they acknowledge that it is open while describing related work as having done most of it.

The models judge both studied and open questions accurately. On the ladders and entity pairs, accuracy is 89 to 99\% when the two groups receive equal weight (Figure~\ref{fig:headline}B). Normal answers \textcolor{black}{still suggest that prior work answers some or all of the question} in 127 of the 210 cases where the separate judgment assigns at most a 10\% probability of prior study (Appendix~\ref{app:search}).

This reporting failure is asymmetric: models hedge on open questions but usually state prior study plainly. For studied ladder questions with the answering paper available, they say so plainly \textcolor{black}{($C=\mathrm{studied}$)} in 88 to 100\% of normal answers across the six systems (97\% pooled, Section~\ref{sec:closest}).

\subsection{Answers overstate what cited papers establish}
\label{sec:claims}

Models overstate how much of a research question a cited paper answers. We grade the coverage claimed in the answer ($a$) and assessed by an independent paper check ($b$) from 0 to 4. Grades 0, 1 and 2 indicate unrelated work, topical overlap and differences in several core parts. Grade 3 means one core part differs, and grade 4 means the paper answers the question. A \emph{near-match claim} has $a\geq3$ and is wrong when $b\leq2$ (Appendix~\ref{app:coding}).

\begin{figure*}[t]
\centering
\includegraphics[width=\textwidth]{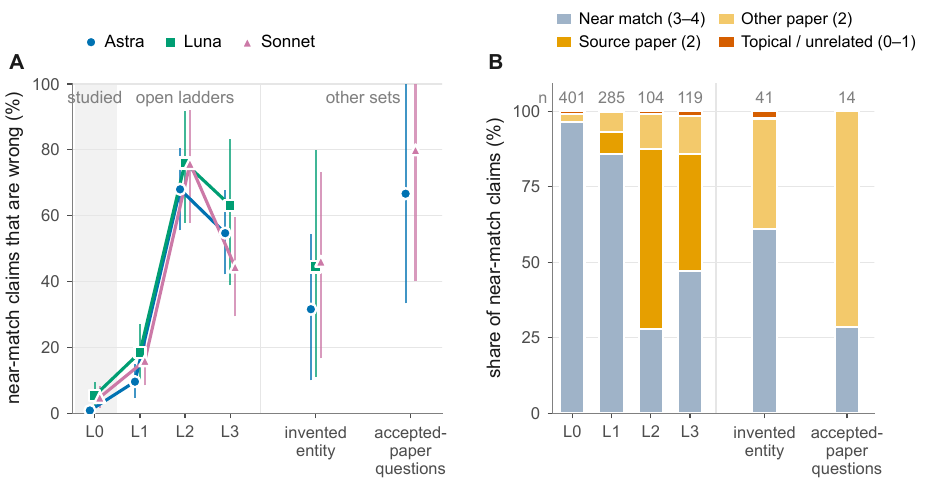}
\caption{\textcolor{black}{How often models' near-match claims are wrong. A compares ladders, invented entities and accepted-paper questions (95\% intervals). On ladders, the error share rises sharply beyond one-part changes. Most near-match claims on accepted-paper questions are wrong. B shows how closely the cited papers actually address the question. Most errors concern papers that differ in several key parts (grade 2), often the ladder\textquotesingle s source paper.}}
\label{fig:nearmatch}
\end{figure*}

The ladder questions let us test whether models recognize how much a proposal differs from published work. Across Astra, Luna and Sonnet, near-match claims are mostly right after one-part changes but mostly wrong after two-part or distant-field changes (Figure~\ref{fig:nearmatch}A). Astra's share of wrong claims rises by 51 points [40, 62] between these groups. Models overstate papers with several differences at similar rates across ladder levels. As genuinely close papers become less common, these errors make up a larger share of near-match claims (Appendix~\ref{app:search}).

Most errors concern the published paper used to build the question: 109 of 138 wrong claims on two-part and distant-field questions. Among papers differing from the question in several core parts, models call that source paper a near match 48\% of the time, compared with 6\% for other papers (Figure~\ref{fig:nearmatch}B).

The models recognize most of their own overstatements when asked to check the papers separately, with full text supplied when available. They reject 120 of the 138 wrong claims on two-part and distant-field questions while retaining 230 of 245 correct claims on one-part questions. Appendix~\ref{app:coding} describes the checks and source-text availability.

\textcolor{black}{Accepted-paper questions} confirm that this overstatement extends beyond the ladders we constructed. Here, 34 of 43 near-match claims are wrong. In separate paper checks, the models reject 31 of those 34 claims.

\subsection{The gap persists when every model reads the same papers}
\label{sec:fixed}

To control for differences in the papers each model retrieves, we repeat the experiment with every model given the same set of papers. All six systems still show reporting gaps. Across systems, separate judgments identify 92\% [88, 96] of the \textcolor{black}{accepted-paper questions} as unstudied, while normal answers state this plainly for 52\% [43, 62]. Almost all remaining answers hedge (Appendix Table~\ref{tab:colm}).

\subsection{Interpreting models' normal answers}
\label{sec:readers}

To simulate how researchers might interpret these answers, we asked other AI models to read the question and answer and decide whether to pursue the project. They could recommend pursuing it as new, pursuing it while explaining how it differs from prior work, or dropping it because it was already done. We refer to these models as \emph{readers}.

Readers were almost four times as likely to recommend explaining what a project adds to prior work \textcolor{black}{when the answer failed to state open status plainly ($C\ne\mathrm{open}$)}: 84\%, compared with 23\% after \textcolor{black}{an $\mathrm{open}$ answer}. For the 30 questions that received \textcolor{black}{answers in both groups} from different systems, we compared reader recommendations question by question. On average, readers recommended explaining what was new 53 percentage points [+40, +66] more often after \textcolor{black}{a $\mathrm{hedged}$ or $\mathrm{studied}$ answer}.

The readers usually assigned a low probability that the question had already been studied and almost never recommended dropping the project (Appendix Table~\ref{tab:reader}). Their recommendations suggest that hedging can make a project's contribution seem less clearly new even when the reader recognizes that its question remains open.

\begin{figure*}[t]
\centering
\includegraphics[width=\textwidth,trim=0 28bp 0 0,clip]{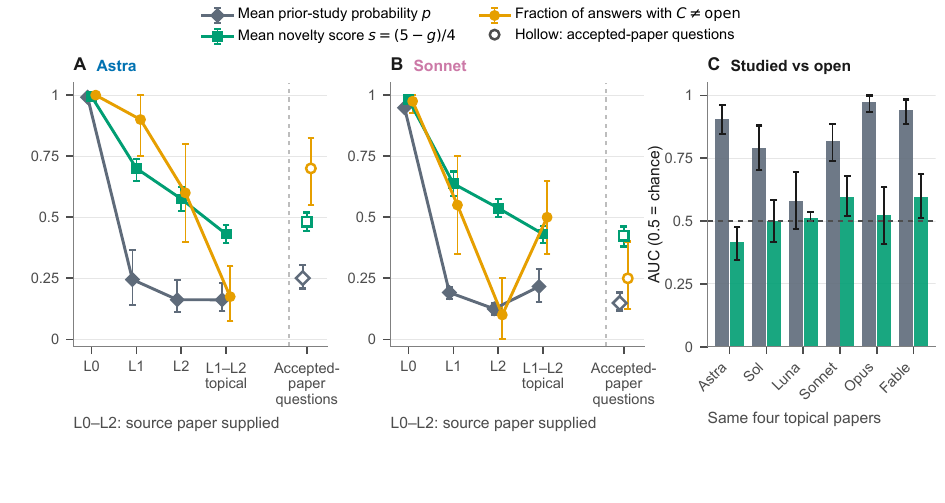}
\caption{\textcolor{black}{With close prior work, models often hedge about questions they judge open. A--B use a 0--1 scale for mean prior-study probability $p$, the fraction of $C\ne\mathrm{open}$ answers and mean novelty score $s$. Higher $s$ means lower novelty. L1--L2 topical replaces the source paper with a topical paper for the same questions. C shows that $p$ distinguishes studied from open questions much better than $s$ (mean AUC 0.84 versus 0.52). Error bars show 95\% intervals.}}
\label{fig:distance}
\end{figure*}

\subsection{The closest paper drives the hedge}
\label{sec:closest}

Close prior work encourages models to hedge about questions they judge to be open. We test this by asking the same open ladder questions twice, replacing only the closest paper with one sharing just the topic and keeping the other three papers unchanged. Across the six systems, \textcolor{black}{the share of $\mathrm{hedged}$ answers} falls from 61\% to 33\%, a drop of 28 points [18, 39]. Their separate judgments identify the questions as open in 97\% of cases with the closest paper present and 91\% after replacement.

Astra shows the contrast clearly. Its hedging falls from 70\% to 18\%, a drop of 52 points [35, 68], while its judgments classify only 5\% of questions as studied in both conditions. The effect is smaller for Anthropic's systems. These systems remember most of the removed papers, which may help explain the smaller effect (Section~\ref{sec:memory}).

The closest paper also affects answers to the questions from newly accepted papers. Removing it reduces hedging by 12 points [3, 20] compared with removing a random, less closely related paper (Appendix Table~\ref{tab:removal}).

\subsection{Remembered papers continue to shape the answer}
\label{sec:memory}

A removed paper can still be available from memory. Across the six systems, replacing it cuts hedging by 47 points when the model does not remember it and 14 when it does, a reduction that is 33 points smaller [16, 50].

We checked memory by asking which work answers the original studied question, without supplying papers. In normal answers with only topical papers supplied, those that name the missing paper hedge 81\% of the time, compared with 17\% of other answers. Removing a close paper from the list therefore need not remove it from the answer.

\begin{figure*}[t]
\centering
\includegraphics[width=\textwidth]{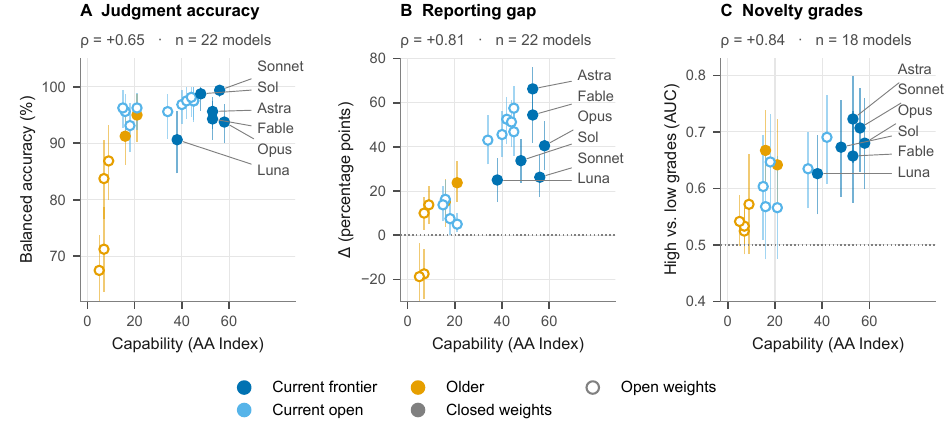}
\caption{\textcolor{black}{More capable models judge study status more accurately (A), yet have larger reporting gaps (B). Balanced accuracy averages the correct judgment rates for studied and open questions. C shows how well novelty grades distinguish high (4--5) from low (1--2) reviewer ratings on RINoBench. Error bars show 95\% intervals. Further comparisons appear in Appendix Figure~\ref{fig:trend_support}.}}
\label{fig:trend}
\end{figure*}

% !TeX root = ../main.tex
\subsection{More capable models judge better but hedge more}
\label{sec:capability}

More capable models give more accurate judgments but have larger reporting gaps. Across the 22 models given the same questions and papers, capability correlates with the gap at +0.81 [+0.55, +0.94] and with judgment accuracy at +0.65 [+0.24, +0.91] (Figure~\ref{fig:trend}A--B, Appendix Table~\ref{tab:trends}). We measure capability using the Artificial Analysis Intelligence Index, a public benchmark score \citep{aa2026index}.

Stronger models also remember more of the source papers (Appendix Figure~\ref{fig:trend_support}C). The association between capability and the gap remains after accounting for the number of papers recalled (Appendix~\ref{app:models}).

\subsection{Using the models' own judgments makes answers clearer}
\label{sec:fix}

A simple correction improves clear reporting in all six systems: build the answer from the model's separate judgment and checks of the cited papers. For Astra, \textcolor{black}{the share of $\mathrm{open}$ answers} rises from 8\% to 95\% on the questions from newly accepted papers, using its original search results. When all systems read the same paper lists, \textcolor{black}{this share} rises from 52\% to 92\%, a gain of 40 points [+32, +48]. It now matches the share of judgments identifying those questions as open. We report the per-system results in Table~\ref{tab:template}.

This \emph{rebuilt answer} uses a template that states the separate judgment first, then explains what the closest paper leaves unanswered, using the model's own paper checks. Other cited papers follow as related work. The template replaces the overstated description of the closest paper and requires no further model call to write the response. Appendix~\ref{app:corrections} compares this approach with rewording and paper checks.

Simply asking the model to state its judgment first produces a smaller gain. The opening agrees with the separate judgment 91\% of the time, but clear reporting rises only from 52\% to 63\%, as later descriptions of prior work still make the question sound partly answered.

Results are mixed when we assess how AI readers interpret the corrected answers. We discuss this further in Appendix~\ref{app:reader_fix}.

\section{\textcolor{black}{Novelty grades reflect closeness to prior work}}
\label{sec:tpr}

So far, we focused on whether a question was studied and how much of it related papers address. Researchers also need to judge how much a project adds to prior work.

\textcolor{black}{We apply TPR's five-point scale \citep{schopf2026think} to our questions to see how novelty grades relate to study status. We then compare predictions of reviewers' grades on RINoBench, which TPR also uses \citep{schopf2026novel}.}

\subsection{\textcolor{black}{Novelty grades and study status}}

\textcolor{black}{On ladders, models give higher novelty grades as the supplied papers cover less of the question (Figure~\ref{fig:distance}A--B).}

\textcolor{black}{We also compare a subset of ladders: 40 studied questions (L0) with their 40 open versions (L1 or L2). Each pair receives the same four topical papers as related work, with the paper that answers the L0 question excluded. We use our judgment prompt to obtain $p=P_{\mathrm{model}}(\mathrm{studied})$ and TPR's grading prompt to obtain a novelty grade $g\in\{1,\ldots,5\}$, rescaled as $s=(5-g)/4$. Across six systems, mean AUC is 52\% for $s$, close to the 50\% chance level, versus 84\% for $p$, a difference of 31 points [26, 37] (Figure~\ref{fig:distance}C). So the models' direct judgments distinguish studied from open questions well, while their novelty grades barely separate the same questions. We describe the experiment in more detail in Appendix~\ref{app:study_auc}.}

\subsection{\textcolor{black}{Comparing model and reviewer grades}}

\textcolor{black}{We next ask whether the models' novelty grades agree with human reviewers, using RINoBench, which pairs research ideas from ICLR 2022 and 2023 submissions with reviewers' novelty grades.}

\textcolor{black}{The ideas reviewers rated most novel are also the ones current models most often call studied. Across six systems, normal answers say this for 58\% of ideas reviewers graded 5, compared with 28\% graded 1 (Figure~\ref{fig:rbmemory}). We set the date in these prompts to the original submission deadline, but models may remember the submission itself. In a separate check without papers or search, our recognition measure flags 72\% of grade-5 ideas, compared with 33\% of grade-1 ideas. These findings challenge RINoBench as a test of novelty at submission: models may already know the paper they are asked to judge.}

\textcolor{black}{This concern applies to our results and TPR's. In our tests of four models used by TPR, the same recognition check flags 3--29\% of ideas. Because recognition is associated with reviewers' grades, a predictor trained on model outputs or internal states could use familiarity with the papers as a clue to those grades (Appendix~\ref{app:rb_memory}). This risk motivates our accepted-paper dataset, which filtered out any work the models may have seen in training (Section~\ref{sec:setup}).}

\textcolor{black}{Novelty grades $g$ separate ideas reviewers rated 4--5 from those rated 1--2 with an AUC of 0.68 across six systems. AUC remains 0.68 without detected recognition and is 0.67 for recognized ideas (Appendix~\ref{app:rb_memory}). These similar AUCs suggest that contamination is less of a concern for this ranking result.}

\textcolor{black}{The normal-answer score $r(C)$ separates the same two groups less well, at AUC 0.40. More capable models' novelty grades rank these ideas better (Figure~\ref{fig:trend}C), with a larger AUC advantage over $r(C)$ (Appendix~\ref{app:rb_memory}).}

\subsection{\textcolor{black}{Predicting reviewer grades from written outputs}}

\textcolor{black}{The ratings help distinguish ideas reviewers rate high or low. We therefore learn from examples graded by reviewers how to turn these ratings into better predictions. TPR trains its predictor on models' internal states, which limits its usage to open-weight models only. Ours uses models' output, so it can work with closed and open models without access to their weights (Appendix~\ref{app:tpr}).}

\textcolor{black}{Our predictor reaches TPR's reported performance range. On the 277 ideas in the RINoBench test set, our predictor scores 24.3 macro-F1, compared with 20.0 to 25.2 for TPR and 17.2 for GPT-5's directly requested grades.}

\section{Conclusion}
\label{sec:conclusion}

\textcolor{black}{As AI use in research and peer review grows, overstated accounts of prior work could discourage useful projects or lead reviewers to undervalue a contribution. Our corrections help models state what remains open and describe the closest paper more accurately. We show how to turn models' written judgments into clearer research advice and predictions of reviewers' novelty grades, without access to their weights.}

\section*{Limitations}
\label{sec:limitations}

Most questions concern machine learning, and open labels rely on peer review and searches that may miss earlier studies. We also rely on AI models to assess answers and cited papers. In the reader experiment, models simulate researchers' interpretations, so effects on actual research decisions remain untested.

As shown, newer models remember some papers in RINoBench data, which could affect both our and TPR's results on the novelty grades experiment.

\bibliography{references}

\appendix
\section{\textcolor{black}{Datasets and Prompts}}
\label{app:sets}

\subsection{\textcolor{black}{Accepted-paper questions}}

{\color{black}
We extracted the accepted-paper questions from the abstracts of the highest-rated COLM 2026 papers with no public version before July 2026. We removed names coined by those papers.

To compare models using the same literature, we give all of them the papers Astra retrieved for each accepted-paper question. Each list contains titles, years and abstracts for 10--33 papers. Independent checks found no answering paper.
\par}

\subsection{\textcolor{black}{Ladders and entity pairs}}

{\color{black}
Table~\ref{tab:controlled_counts} summarizes the full set and the 100-question subset used to compare normal answers with separate judgments. Tests with supplied papers use 40 ladders, each with a studied question and one open version: 20 L1 and 20 L2.
\par}

\begin{table}[htbp]
\centering
\small
\setlength{\tabcolsep}{3pt}
\begin{tabular}{@{}llrr@{}}
\toprule
Question type & Status & Full set & Subset \\
\midrule
L0: Published question & Studied & 108 & 20 \\
L1: One part changed & Open & 107 & 20 \\
L2: Two parts changed & Open & 108 & 15 \\
L3: Distant field & Open & 108 & 15 \\
Real entity & Studied & 36 & 12 \\
Invented entity & Open & 36 & 18 \\
\midrule
All studied & & 144 & 32 \\
All open & & 359 & 68 \\
Total & & 503 & 100 \\
\bottomrule
\end{tabular}
\caption{\textcolor{black}{Question counts for ladders and entity pairs. L0 is studied, while L1--L3 are the retained open versions.}}
\label{tab:controlled_counts}
\end{table}

\subsection{\textcolor{black}{RINoBench}}

{\color{black}
We evaluate the grading predictor on all 277 RINoBench test ideas, which have reviewers' novelty-and-significance grades from 1 to 5 \citep{schopf2026novel}.

The recognition and ranking analyses use a separate sample of 150 ideas, with 30 per grade. It combines 135 test ideas with 15 grade-1 training ideas because the test set contains only 15 grade-1 ideas. These added training ideas are excluded from the predictor's test results.
\par}

\subsection{\textcolor{black}{Prompts and model settings}}
\label{app:prompts}

{\color{black}
Each response starts in a fresh session. The normal request begins: ``I'm about to start a research project and want to check the literature first. Has the following research question already been studied?'' The forced request asks for YES or NO, the probability of prior study from 0 to 100, and an answering work after YES.

When papers are supplied, normal answers and judgments receive the same list without search tools. In live-search tests, they use the same retrieved papers for accepted-paper questions and search independently for ladders and entity pairs.

The frontier experiments use medium reasoning effort for OpenAI and default effort for Anthropic. The broader model comparison uses each provider's default reasoning setting. Table~\ref{tab:model_list} lists the 22 models with complete results. Capability is measured by the Artificial Analysis Intelligence Index v4.3.2 \citep{aa2026index}.
\par}

\begin{table*}[!t]
\centering
\small
\begin{tabular}{p{0.19\textwidth}p{0.74\textwidth}}
\toprule
Group & Models \\
\midrule
Current frontier & GPT-6-Astra, GPT-6-Sol, GPT-6-Luna, Claude Sonnet 5.5, Claude Opus 5.5, Claude Fable 5.1 \\
Current open weights & Qwen3.8 27B, Qwen3.8 Flash, Qwen3.8 2.4T, GLM 5.3 Flash, GLM 5.3, Kimi K3, Qwen3.5 122B, Gemma 4 31B, Qwen3.6 27B, Qwen3.6 35B-A3B \\
Older & Gemini 2.5 Pro, GPT-5 mini, gpt-oss-20b, Qwen3 32B, Llama 3.1 70B, Gemma 3 27B \\
\bottomrule
\end{tabular}
\caption{\textcolor{black}{Models tested with both answer formats on the same 120 questions and paper lists.}}
\label{tab:model_list}
\end{table*}

\section{\textcolor{black}{Answer Labels and Paper Checks}}
\label{app:coding}

{\color{black}
Two model coders, one from each frontier lab, classify the entire normal answer using $C\in\{\mathrm{studied},\mathrm{hedged},\mathrm{open}\}$ from Section~\ref{sec:setup}. They see neither the system identity nor the true study status. A model from the other lab than the answering system settles disagreements.

For each cited paper, coders record how closely the answer claims it addresses the question ($a$). An independent grader, GPT-6-Astra, assesses the paper\textquotesingle s coverage ($b$) without seeing that answer. Both use Table~\ref{tab:rubric}. A near-match claim has $a\geq3$ and is wrong when $b\leq2$. The answering systems then check their own claims separately. Full text was available for 538 of 551 checks. Excluding Astra's answers, its fellow systems still reject 25 of 28 wrong claims on live-search accepted-paper questions.
\par}

\begin{table}[htbp]
\centering
\small
\begin{tabular}{cp{0.77\columnwidth}}
\toprule
Grade & Coverage of the research question \\
\midrule
0 & Unrelated \\
1 & Topical \\
2 & Differs in several core parts \\
3 & Differs in one core part \\
4 & Answers the question \\
\bottomrule
\end{tabular}
\caption{\textcolor{black}{Paper coverage, assessed against the question\textquotesingle s required parts. Claimed, independent and self-assessed grades use the same scale.}}
\label{tab:rubric}
\end{table}

{\color{black}
Before disagreements are resolved, answer-label agreement is 87\% across 1,215 normal answers ($\kappa=0.80$), ranging from 78\% on accepted-paper questions with supplied papers to 88\% on RINoBench. Agreement on near-match claims is 94--95\% ($\kappa=0.54$--$0.81$). These checks measure agreement among models.
\par}

\section{\textcolor{black}{Reporting Gaps and Paper Claims}}
\label{app:search}

\subsection{\textcolor{black}{Answers with live search}}

{\color{black}
For the low-probability comparison ($p\leq0.1$), we count claims of partial or complete prior study and exclude answers that merely leave status unclear.
\par}

\begin{table*}[!t]
\centering
\small
\begin{tabular}{llrrrr}
\toprule
Question set & System & $C=\mathrm{open}$ (\%) & $J=\mathrm{NO}$ (\%) & $\Delta$ (points) & Balanced judgment accuracy (\%) \\
\midrule
\shortstack[l]{Accepted-paper\\questions} & Astra & 8 [0, 18] & 98 & +90 [80, 98] & N/A \\
 & Luna & 70 [55, 82] & 100 & +30 & N/A \\
 & Sol & 38 [22, 52] & 100 & +62 & N/A \\
 & Sonnet & 68 [52, 82] & 98 & +30 & N/A \\
 & Opus & 48 [32, 62] & 78 & +30 & N/A \\
 & Fable & 35 [20, 50] & 92 & +57 & N/A \\
\midrule
\shortstack[l]{Ladders and entity\\pairs (subset)} & Astra & 12 [5, 19] & 99 & +87 [78, 94] & 96 \\
 & Luna & 53 [42, 65] & 94 & +41 [29, 54] & 89 \\
 & Sol & 35 [25, 46] & 100 & +65 [54, 75] & 97 \\
 & Sonnet & 44 [34, 55] & 93 & +49 [36, 60] & 95 \\
 & Opus & 27 [16, 37] & 91 & +64 [52, 76] & 95 \\
 & Fable & 44 [31, 55] & 98 & +55 [43, 68] & 99 \\
\bottomrule
\end{tabular}
\caption{\textcolor{black}{Normal answers and separate judgments with live search. The reporting gap $\Delta$ is measured on open questions. Balanced judgment accuracy also includes studied questions.}}
\label{tab:verdicts}
\end{table*}

\subsection{\textcolor{black}{Answers with shared papers}}
\label{app:colm}

{\color{black}
To test whether the reporting gap persists without live search, we give all six systems the papers Astra found for each accepted-paper question (Table~\ref{tab:colm}). Astra's share of $\mathrm{open}$ answers rises from 8\% when it searches to 30\% when we supply the same papers.
\par}

\begin{table*}[!t]
\centering
\small
\setlength{\tabcolsep}{4pt}
\begin{tabular}{lrrrrrrr}
\toprule
System & \shortstack{$C=\mathrm{open}$\\(\%)} & \shortstack{$C=\mathrm{hedged}$\\(\%)} & \shortstack{$C=\mathrm{studied}$\\(\%)} & $J=\mathrm{NO}$ (\%) & Mean $p$ (\%) & $\Delta$ (points) & \shortstack{Rebuilt\\$C=\mathrm{open}$ (\%)} \\
\midrule
Astra & 30 [18, 45] & 68 & 2 & 95 [88, 100] & 25 & +65 [+50, +80]  & 95 \\
Sol & 57 [42, 72] & 42 & 0 & 100 [100, 100] & 8 & +43 [+28, +57]  & 100 \\
Luna & 72 [57, 85] & 28 & 0 & 95 [88, 100] & 21 & +22 [+10, +38]  & 95 \\
Sonnet & 75 [62, 88] & 25 & 0 & 98 [92, 100] & 15 & +22 [+10, +35]  & 98 \\
Opus & 42 [28, 57] & 57 & 0 & 85 [72, 95] & 34 & +42 [+28, +57]  & 85 \\
Fable & 38 [22, 52] & 62 & 0 & 82 [70, 92] & 32 & +45 [+27, +62]  & 82 \\
\midrule
Pooled & 52 [43, 62] & 47 & 0 & 92 [88, 96] & 22 & +40 [+31, +48]  & 92 \\
\bottomrule
\end{tabular}
\caption{\textcolor{black}{Normal answers, separate judgments and corrections on accepted-paper questions, with the same papers supplied to every system. Rebuilt answers use the correction in Section~\ref{sec:fix}. Rates are percentages and gaps are percentage points.}}
\label{tab:colm}
\label{tab:template}
\end{table*}

{\color{black}
Independent checks reject 26 of 38 near-match claims about supplied papers, or 68\% [38, 100]. The systems' own checks reject 21 of those 26 wrong claims and retain 9 of the 12 correct claims.
\par}

\subsection{\textcolor{black}{Which paper claims are wrong}}

{\color{black}
Of the 138 wrong near-match claims on two-part and distant-field questions, 135 concern grade-2 papers. Across ladder levels, Astra, Luna and Sonnet describe 18--24\% of grade-2 citations as near matches. As genuinely close papers become less common, these overstatements make up a larger share of claims (Figure~\ref{fig:nearmatch}).

Depending on the system, 58--90\% of these errors accompany an acknowledgment that the question is open.
\par}

\subsection{\textcolor{black}{How readers interpret the answers}}

{\color{black}
To compare advice about the same project, we use the 30 questions with both $C=\mathrm{open}$ and $C\ne\mathrm{open}$ answers, weighting each question equally. Table~\ref{tab:reader} summarizes readers' recommendations.
\par}

\begin{table}[htbp]
\centering\small
\setlength{\tabcolsep}{3pt}
\begin{tabular}{@{}lrr@{}}
\toprule
System & \shortstack{Pursue, explain\\contribution (\%)} & \shortstack{Reader minus model\\probability (points)} \\
\midrule
Astra & 68 & +2 [$-1$, +6] \\
Sol & 40 & +12 [+9, +15] \\
Luna & 22 & $-2$ [$-12$, +6] \\
Sonnet & 42 & +10 [+7, +13] \\
Opus & 72 & $-0$ [$-6$, +6] \\
Fable & 68 & +0 [$-5$, +6] \\
\midrule
Pooled & 52 & +4 [+1, +6] \\
\bottomrule
\end{tabular}
\caption{\textcolor{black}{Reader advice and differences in estimated prior-study probability. Sonnet reads OpenAI answers and Luna reads Anthropic answers. Of 240 answers, only one prompts a recommendation to drop the project. Other recommendations are to pursue it as new.}}
\label{tab:reader}
\end{table}

\section{\textcolor{black}{Close Papers and Memory}}
\label{app:ladders}

\subsection{\textcolor{black}{Replacing a ladder's source paper}}

{\color{black}
Each ladder question receives its source paper and three topical papers. We replace only the source paper with another topical paper. Independent checks rate the topical papers 0 or 1. The 22-model comparison uses the initial lists. Table~\ref{tab:ladders} compares the paper conditions for Astra and Sonnet.

Replacing the source paper reduces hedging by about 50 points for each OpenAI system. The change is smaller for Anthropic's systems: hedging falls by 12 points for Sonnet and 5 for Fable, and rises by 2 for Opus.

For Astra, the reduction is larger on one-part questions (80 points [60, 95]) than on two-part questions (25 points [5, 50]).
\par}

\begin{table*}[!t]
\centering
\small
\setlength{\tabcolsep}{3.5pt}
\begin{tabular}{llrrrrr}
\toprule
System & Question and list & $C=\mathrm{studied}$ & $C=\mathrm{hedged}$ & $C=\mathrm{open}$ & $J=\mathrm{YES}$ & Novelty grade $g$ \\
\midrule
Astra & Studied, answering paper listed & 100 & 0 & 0 & 100 & 1.10 \\
 & Open, source paper listed & 0 & 70 & 30 & 5 & 2.55 \\
 & Open, topical only & 0 & 18 & 82 & 5 & 3.27 \\
 & Studied, answering paper absent & 45 & 12 & 42 & 68 & 3.42 \\
\midrule
Sonnet & Studied, answering paper listed & 98 & 2 & 0 & 98 & 1.05 \\
 & Open, source paper listed & 0 & 62 & 38 & 0 & 2.62 \\
 & Open, topical only & 0 & 50 & 50 & 15 & 3.27 \\
 & Studied, answering paper absent & 68 & 18 & 15 & 55 & 2.98 \\
\bottomrule
\end{tabular}
\caption{\textcolor{black}{Forty ladders under each paper condition. Answer-label and judgment columns are percentages. The final column is the mean novelty grade.}}
\label{tab:ladders}
\end{table*}

\subsection{\textcolor{black}{Removing a paper from accepted-paper lists}}

{\color{black}
Across the six systems, 53.8\% of answers hedge with the full list and 54.6\% after removing a random, less-related paper. Removing the closest paper shortens the list by the same amount and lowers hedging to 42.9\% (Table~\ref{tab:removal}). The full-list rate was 6.7 points [2.1, 11.7] higher on this run than on the first run of the same prompt.
\par}

\begin{table*}[!t]
\centering
\small
\begin{tabular}{lrrrr}
\toprule
 & \multicolumn{3}{c}{$C=\mathrm{hedged}$ (\%)} & Closest minus random \\
\cmidrule(lr){2-4}\cmidrule(lr){5-5}
System & \shortstack[r]{All\\papers} & \shortstack[r]{Closest\\removed} & \shortstack[r]{Random\\removed} & removal (points) \\
\midrule
Astra & 72.5 & 52.5 & 70.0 & $-17.5$ [$-32.5$, $-2.5$] \\
Sol & 47.5 & 32.5 & 60.0 & $-27.5$ [$-42.5$, $-15.0$] \\
Luna & 35.0 & 37.5 & 40.0 & $-2.5$ [$-20.0$, +15.0] \\
Sonnet & 32.5 & 22.5 & 35.0 & $-12.5$ [$-27.5$, +2.5] \\
Opus & 60.0 & 47.5 & 52.5 & $-5.0$ [$-20.0$, +10.0] \\
Fable & 75.0 & 65.0 & 70.0 & $-5.0$ [$-22.5$, +10.0] \\
\midrule
Pooled & 53.8 & 42.9 & 54.6 & $-11.7$ [$-20.4$, $-3.3$] \\
\bottomrule
\end{tabular}
\caption{\textcolor{black}{Hedged answers ($C=\mathrm{hedged}$) after removing one paper from accepted-paper lists. Each system answers 40 questions. Rates are percentages, and differences include 95\% intervals.}}
\label{tab:removal}
\end{table*}

\subsection{\textcolor{black}{Recalling an absent paper}}

{\color{black}
Models often recognize a studied question even when its answering paper is absent. Table~\ref{tab:memory} groups answers by whether the model named that paper in a separate recall check, without supplied papers or search.
\par}

\begin{table}[htbp]
\centering\small
\setlength{\tabcolsep}{3pt}
\begin{tabular}{@{}llrrr@{}}
\toprule
System & Paper & $n$ & $C=\mathrm{studied}$ & $J=\mathrm{YES}$ \\
\midrule
Astra & Supplied & 40 & 100 & 100 \\
 & Absent, recalled & 26 & 62 & 92 \\
 & Not recalled & 14 & 14 & 21 \\
Sonnet & Supplied & 40 & 98 & 98 \\
 & Absent, recalled & 28 & 89 & 68 \\
 & Not recalled & 12 & 17 & 25 \\
\bottomrule
\end{tabular}
\caption{\textcolor{black}{Studied ladders with the answering paper supplied or absent. Absent papers are grouped by separate recall. $C=\mathrm{studied}$ and $J=\mathrm{YES}$ are percentages.}}
\label{tab:memory}
\end{table}

\subsection{\textcolor{black}{Plausibility and distance}}

\begin{figure*}[t]
\centering
\includegraphics[width=\textwidth]{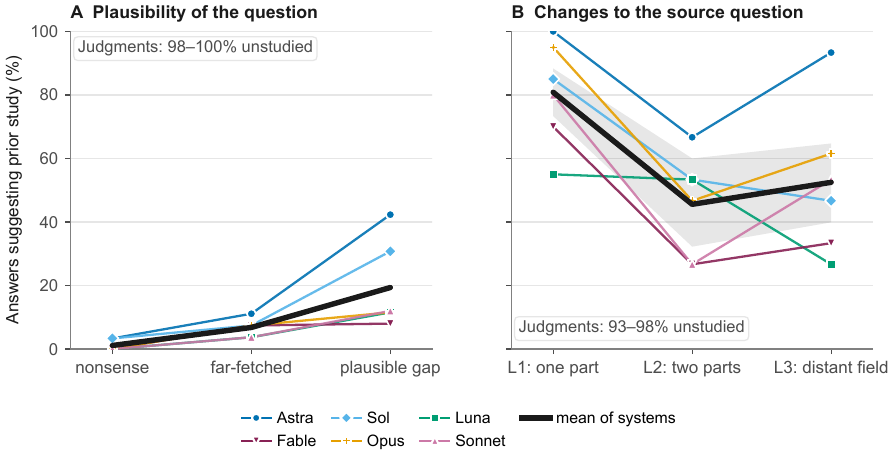}
\caption{\textcolor{black}{Answers implying partial or complete prior study become more common as questions sound more plausible (A) and when only one part of a published question changes (B). The measure excludes answers that are merely unclear. Separate judgments usually classify these questions as unstudied.}}
\label{fig:search_triggers}
\end{figure*}

{\color{black}
The plausibility gradient combines real terms in a fixed question template. How often the terms appear together in published papers defines three levels: nonsense (30 questions), far-fetched (27) and plausible gaps (26). All are open questions. Controls include 15 shuffled questions and 15 studied questions.

Figure~\ref{fig:search_triggers} counts answers that imply partial or complete prior study, excluding answers that merely leave status unclear. This share rises from 1\% to 7\% to 19\% across the three levels, averaging +9 points [5, 14] per level. Separate judgments identify 98--100\% as unstudied.

On live-search ladders, the same measure is 81\% [73, 88] after one-part changes, 46\% [32, 60] after two-part changes and 52\% [40, 65] after distant-field changes. Separate judgments identify 93--98\% as open. A distant-field question can still have close work elsewhere in the literature.
\par}

\section{\textcolor{black}{Capability and Reporting}}
\label{app:models}

\begin{figure*}[t]
\centering
\includegraphics[width=\textwidth]{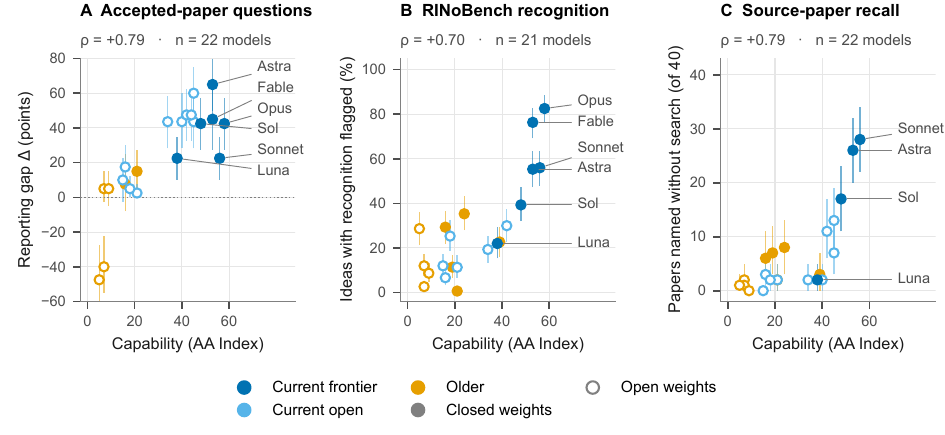}
\caption{\textcolor{black}{Additional comparisons with model capability. A shows the reporting gap on accepted-paper questions, B recognition of RINoBench ideas, and C recall of ladder source papers. Error bars show 95\% intervals.}}
\label{fig:trend_support}
\end{figure*}

{\color{black}
Current models have balanced judgment accuracy of 91--99\%, compared with 68--99\% across all models. Some older models wrongly classify open questions as studied, producing negative reporting gaps. We therefore compare judgment accuracy alongside the gap (Table~\ref{tab:trends}).
\par}

\begin{table*}[t]
\centering
\small
\begin{tabular}{lrr}
\toprule
Measure against capability & Models & Spearman correlation \\
\midrule
Gap, combined open questions & 22 & +0.81 [+0.55, +0.94] \\
Gap, accepted-paper questions & 22 & +0.79 [+0.50, +0.92] \\
$C=\mathrm{open}$ share, accepted-paper questions & 22 & $-0.75$ [$-0.90$, $-0.48$] \\
Balanced judgment accuracy, complete set & 22 & +0.65 [+0.24, +0.91] \\
Balanced judgment accuracy, ladders & 25 & +0.68 [+0.33, +0.89] \\
Gap, open ladders & 25 & +0.73 [+0.51, +0.86] \\
Gap, combined open questions, current models & 16 & +0.59 [+0.07, +0.89] \\
\bottomrule
\end{tabular}
\caption{\textcolor{black}{Spearman correlations with capability. Confidence intervals resample models. The 22-model results use the complete comparison, while the 25-model results also include ladder-only runs.}}
\label{tab:trends}
\end{table*}

{\color{black}
The gap's correlation with capability remains +0.78 [0.59, 0.91] after accounting for source-paper recall. After accounting for capability, its correlation with recall is $-0.30$ [$-0.74$, 0.37].
\par}

\section{\textcolor{black}{Correcting Answers}}
\label{app:corrections}

{\color{black}
Rebuilt answers state the model's separate judgment, so they preserve errors in that judgment (Table~\ref{tab:template}).
\par}

\subsection{\textcolor{black}{Rewording and reading the cited papers}}

{\color{black}
We compare the template with a reworded request, ``Which published studies have addressed the following research question, and what did they find?'' We also compare checking cited papers through abstracts or full text. Full-text checks revisit papers that may answer the question, including related papers when a YES judgment lacks support.

Table~\ref{tab:search_fix} compares these approaches on the same 30 studied and 30 open ladder and entity-pair questions. Reading full texts raises Astra's retention of answering papers from 73\% to 100\%. Sonnet retains all of them with either approach.
\par}

\begin{table*}[!t]
\centering
\small
\begin{tabular}{llrrrrr}
\toprule
Measure & System & Original & Template & Reworded & Abstracts & Full text \\
\midrule
$C=\mathrm{open}$ (\%) & Astra & 3 & 83 & 80 & 57 & 90 [79, 100] \\
 & Sonnet & 47 & 93 & 90 & 50 & 87 [72, 97] \\
True status plainly (\%) & Astra & 47 [35, 59] & 88 [80, 95] & 85 & N/A & 90 [83, 97] \\
 & Sonnet & 68 & 95 [89, 100] & 90 & N/A & 86 \\
Answering paper retained (\%) & Astra & 100 & 93 & 100 & 73 & 100 \\
 & Sonnet & 97 & 97 & 100 & 100 & 100 \\
\bottomrule
\end{tabular}
\caption{\textcolor{black}{Corrections on the same 60 ladder and entity-pair questions with live search. $C=\mathrm{open}$ rates use the 30 open questions. True-status accuracy weights studied and open questions equally. Retention measures whether an answering paper remains in the answer.}}
\label{tab:search_fix}
\end{table*}

{\color{black}
On accepted-paper questions, checking full texts instead of abstracts raises the share of $\mathrm{open}$ answers from 62\% to 92\% for Astra and from 78\% to 90\% for Sonnet.

Asking ``Is this research question novel?'' produces the largest gaps among the alternative requests: 98 points for Astra and 82 for Sonnet.
\par}

\subsection{\textcolor{black}{How readers respond to the correction}}
\label{app:reader_fix}

{\color{black}
After correction, Sonnet and Luna less often recommend explaining what the project adds when the original answer was $\mathrm{hedged}$ or $\mathrm{studied}$, and more often when it was $\mathrm{open}$. The two shifts largely cancel (52\% to 49\% overall). Opus and Astra make that recommendation less often in both groups (Table~\ref{tab:reader_correction}).
\par}

\begin{table*}[!t]
\centering\small
\begin{tabular}{llrrr}
\toprule
Readers & Original answer group & Normal (\%) & Rebuilt (\%) & Change (points) \\
\midrule
Sonnet / Luna & $C\ne\mathrm{open}$ & 84 & 58 & $-26$ [$-40$, $-11$] \\
 & $C=\mathrm{open}$ & 23 & 40 & +17 [6, 27] \\
 & All & 52 & 49 & $-3$ [$-14$, +7] \\
\midrule
Opus / Astra & $C\ne\mathrm{open}$ & 96 & 50 & $-46$ \\
 & $C=\mathrm{open}$ & 46 & 29 & $-17$ \\
 & All & 70 & 39 & $-31$ [$-40$, $-22$] \\
\bottomrule
\end{tabular}
\caption{\textcolor{black}{Recommendations to pursue while explaining the contribution, before and after correction. The first reader in each pair reads OpenAI answers and the second reads Anthropic answers. The groups contain 114 non-$\mathrm{open}$ and 126 $\mathrm{open}$ answers.}}
\label{tab:reader_correction}
\end{table*}

\section{\textcolor{black}{Novelty Grades and Reviewer Predictions}}
\label{app:tpr}

\subsection{\textcolor{black}{Study status and novelty grades}}
\label{app:study_auc}

{\color{black}
On ladders, Astra's and Sonnet's mean novelty grades rise from 1.0 with an answering paper (L0), to 2.2--2.5 with one part different (L1), 2.7--2.9 with two parts different (L2), and 3.3 with topical papers only (Figure~\ref{fig:distance}A--B).

The ladder comparison in Section~\ref{sec:tpr} uses prior-study probability $p$ and the rescaled novelty grade $s=(5-g)/4$.

For a score $s$, let $s_{\mathrm{studied}}$ and $s_{\mathrm{open}}$ denote scores of independently sampled studied and open questions.
\[
\begin{aligned}
\operatorname{AUC}(s)={}&\Pr(s_{\mathrm{studied}}>s_{\mathrm{open}})\\
&+\tfrac12\Pr(s_{\mathrm{studied}}=s_{\mathrm{open}}).
\end{aligned}
\]
Every studied--open pair contributes, with half credit for ties. Reported AUCs are averaged across six systems.

Lower novelty grades also accompany less clear reporting of open status. All accepted-paper questions are open, yet models state this plainly in 9\% of answers at grade 2, compared with 54\% at grade 3 and 73\% at grade 4.
\par}

\subsection{\textcolor{black}{Ranking RINoBench ideas}}
\label{app:rb_memory}

{\color{black}
Novelty grades separate reviewer ratings of 4--5 from 1--2 with an AUC of 0.68 [0.61, 0.74] across the six frontier systems. AUC is 0.67 with detected recognition and 0.68 without it, compared with 0.40 for $r(C)$ overall.

On paired comparisons, novelty grades outperform $r(C)$ by 0.27 AUC [0.20, 0.34]. Across 18 models, capability correlates with novelty-grade AUC at +0.84 [0.58, 0.93] and with its advantage over $r(C)$ at +0.91 [0.67, 0.99].
\par}

\subsection{\textcolor{black}{Recognition of evaluated submissions}}

\begin{figure}[!htbp]
\centering
\includegraphics[width=\columnwidth,trim=0 10bp 0 0,clip]{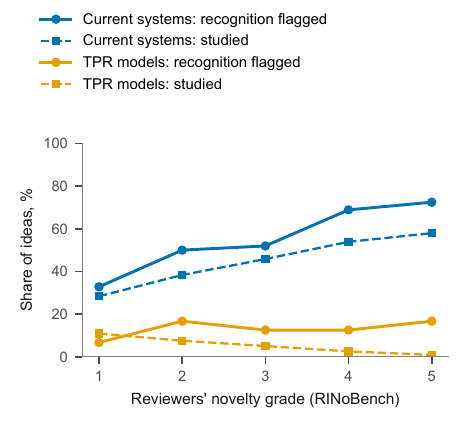}
\caption{\textcolor{black}{Current models more often recognize highly rated RINoBench ideas and label them $C=\mathrm{studied}$. Each grade contains 30 ideas. Recognition is lower for the four TPR models.}}
\label{fig:rbmemory}
\end{figure}

{\color{black}
Recognition checks ask for an answering work's title and authors, without supplied papers or search. Both these checks and normal answers use the original ICLR abstract deadline as the reference date.

We flag recognition when a citation is absent from the idea's references and dated in the deadline year or later. For models that usually omit years, we match their titles to dated citations from other models. This check can also flag different or incorrectly named work. It does not measure recall of reviewers' grades.

Recognition rates are 3--29\% for the four TPR models (13\% pooled), compared with 55\% for the six current systems (Figure~\ref{fig:rbmemory}). Normal answers classify recognized ideas as studied 16 points [9, 22] more often, after accounting for reviewer grade, stated prior-study probability and differences between systems. These checks do not isolate memory's contribution to TPR's published scores.
\par}

\subsection{\textcolor{black}{Predicting all five reviewer grades}}
\textcolor{black}{Models often assign middle grades to ideas reviewers rated low. Astra and Sonnet assign grade 3 in 48 of 60 ratings of ideas reviewers graded 1 or 2. Similar behavior appears in TPR \citep{schopf2026think} and LLM grading of student assignments \citep{golchin2025grading}. TPR's rubric may encourage this: grade 3 includes incremental updates and new combinations of known approaches.}

{\color{black}
We tested whether a missing reference could explain the middle grades. For a separate sample of 20 ICLR submissions rated low in novelty, we added a close paper ($b=3$) that a reviewer cited and the authors had omitted. Only three of Astra and Sonnet's 40 ratings changed, all from 3 to 2.

On the 277 RINoBench test ideas, directly requested grades score 16.7--20.3 macro-F1 across the six systems, averaging 18.4 [15.4, 21.3]. Macro-F1 averages F1 across the five grades, reported on a 0--100 scale. No current system assigns grade 5.
\par}

{\color{black}
Our predictor combines six models' written grades and three expected grades calculated from their stated probabilities. We put these nine estimates on the same scale and average them. Four cutoffs learned from 215 training ideas convert the average to a grade from 1 to 5. Adding the probability-based estimates raises test macro-F1 from 21.8 to 24.3.

Using the same training ideas, we fit a predictor to each TPR model's repeated grades or grade probabilities. TPR scores higher for all four (Table~\ref{tab:tpr_same}).
\par}

\medskip\noindent\begin{minipage}{\columnwidth}
\centering\small
\setlength{\tabcolsep}{3pt}
\begin{tabular}{@{}lrr@{}}
\toprule
Model & Written outputs & TPR \\
\midrule
Gemma 3 27B & 12.7 [9.1, 16.9] & 25.2 \\
gpt-oss-20b & 15.7 [11.9, 19.5] & 22.8 \\
Qwen3 32B & 21.0 [15.8, 26.4] & 21.1 \\
Llama 3.1 70B & 19.2 [14.5, 24.2] & 20.6 \\
\bottomrule
\end{tabular}
\captionof{table}{\textcolor{black}{RINoBench test macro-F1 using each model's written outputs (ours) or internal states (TPR).}}
\label{tab:tpr_same}
\end{minipage}\par

\end{document}